\documentclass{article}

\usepackage{PRIMEarxiv}

\usepackage[utf8]{inputenc} % allow utf-8 input
\usepackage[T1]{fontenc}    % use 8-bit T1 fonts
\usepackage{hyperref}       % hyperlinks
\usepackage{url}            % simple URL typesetting
\usepackage{booktabs}       % professional-quality tables
\usepackage{amsfonts}       % blackboard math symbols
\usepackage{nicefrac}       % compact symbols for 1/2, etc.
\usepackage{microtype}      % microtypography
\usepackage{lipsum}
\usepackage{float}
\usepackage{fancyhdr}       % header
\usepackage{graphicx}       % graphics
\graphicspath{{media/}}     % organize your images and other figures under media/ folder
\usepackage{subcaption}
\usepackage{caption}
\usepackage{tikz}
\usepackage{pgfplots}
\pgfplotsset{compat=1.18} 
\title{NagaNLP: Bootstrapping NLP for Low-Resource Nagamese Creole with Human-in-the-Loop Synthetic Data}

\author{
  Agniva Maiti \\
  RespAI Lab \\
  KIIT Bhubaneswar \\
  Bhubaneswar, India \\
  \texttt{maitiagniva@gmail.com}
  \and
  Manya Pandey \\
  RespAI Lab \\
  KIIT Bhubaneswar \\
  Bhubaneswar, India \\
  \texttt{manyapandey7842@gmail.com}
  \and
  Murari Mandal\thanks{Corresponding author: \texttt{murari.mandalfcs@kiit.ac.in
}} \\
  RespAI Lab \\
  KIIT Bhubaneswar \\
  Bhubaneswar, India \\
  \texttt{murari.mandalfcs@kiit.ac.in
}
}

\begin{document}
\maketitle

\begin{abstract}
The vast majority of the world's languages, particularly creoles like Nagamese, remain severely under-resourced in Natural Language Processing (NLP), creating a significant barrier to their representation in digital technology. This paper introduces NagaNLP, a comprehensive open-source toolkit for Nagamese, bootstrapped through a novel methodology that relies on LLM-driven but human-validated synthetic data generation. We detail a multi-stage pipeline where an expert-guided LLM (Gemini) generates a candidate corpus, which is then refined and annotated by native speakers. This synthetic-hybrid approach yielded a 10K pair conversational dataset and a high-quality annotated corpus for foundational tasks.
To assess the effectiveness of our methodology, we trained both discriminative and generative models. Our fine-tuned XLM-RoBERTa-base model establishes a new benchmark for Nagamese, achieving a 93.81\% accuracy (0.90 F1-Macro) on Part-of-Speech tagging and a 0.75 F1-Macro on Named Entity Recognition, massively outperforming strong zero-shot baselines. Furthermore, we fine-tuned a Llama-3.2-3B Instruct model, named NagaLLaMA, which demonstrates superior performance on conversational tasks, achieving a Perplexity of 3.85, an order of magnitude improvement over its few-shot counterpart (96.76). We release the NagaNLP toolkit, including all datasets, models, and code, providing a foundational resource for a previously underserved language and a reproducible framework for reducing data scarcity in other low-resource contexts.
\end{abstract}


\section{Introduction}

Language technologies are pivotal in promoting multilingualism and preserving the world's linguistic diversity. However, of the over 7,000 languages spoken today, only a small fraction are represented in the rapidly advancing landscape of Natural Language Processing (NLP), creating a ``digital cliff'' for under-resourced communities \cite{JoshiEtAl2020, Kornai2013,jayan2025challenges,shafer2025building}. This disparity is particularly acute for creole languages, which, despite serving vibrant communities, often lack the standardized corpora and institutional support necessary for digital integration.

This paper addresses the severe lack of data for Nagamese Creole, an Assamese-lexified lingua franca  spoken across Nagaland in Northeast India. While Assamese has seen recent resource development\cite{ BaruahSarma2023, DasSingh2024, BaruahHazarika2014, ChoudhuryEtAl2024}, Nagamese, being an oral vernacular with limited digital footprint, has remained largely inaccessible to modern NLP. The only notable prior work involves a Part-of-Speech (POS) tagger built using Conditional Random Fields (CRF) on a small, manually created corpus by \cite{ShoheEtAl2025}, highlighting the foundational resource gap. This typifies the classic ``chicken-and-egg'' problem for low-resource languages: without data, no models can be built, and without models, the cost and effort of creating high-quality data from scratch are often prohibitive\cite{MagueresseEtAl2020, VerdonikEtAl2025}.

To break this cycle, we propose a novel and scalable ``LLM-to-human'' bootstrapping pipeline. Our methodology leverages a state-of-the-art Large Language Model (LLM) as a ``language elicitor,'' guided by expert human interaction, to generate a large-scale, high-quality synthetic corpus. This raw data is then meticulously validated, corrected, and annotated by native speakers, transforming it into a reliable resource. We hypothesize that this human-validated, synthetic-hybrid corpus can effectively bootstrap an entire NLP ecosystem for an extremely low-resource language.

We validate this hypothesis by using the generated data to build the first comprehensive NLP toolkit for Nagamese, which we call \textit{NagaNLP}. Our contributions are threefold:
\begin{enumerate}
    \item We present a novel, replicable bootstrapping methodology that efficiently bridges the data gap for extremely low-resource languages.
    \item We release \textit{NagaNLP}, the first open-source toolkit for Nagamese Creole, which includes: (a) a human-validated, annotated corpus for POS tagging and Named Entity Recognition (NER), and (b) a larger conversational corpus for generative tasks.
    \item We provide strong empirical validation of our methodology by training foundational NLP models that significantly outperform existing baselines and successfully instruction-tuning a state-of-the-art generative model, Llama 3.2, on our synthetic-hybrid data for complex downstream tasks.
\end{enumerate}

\section{Related Work}

Our research is situated at the intersection of four key areas in NLP: low-resource language processing~\cite{mujadia2025ilgov,abdu2025empirical,chakravarthi2025detecting}, synthetic data generation, NLP for creole languages, and the adaptation of Large Language Models. While approaches like \cite{SchickSchuetze2021} focus on generating labeled data for existing tasks, our pipeline is designed for the zero-resource setting\cite{LampleEtAl2018} where linguistic knowledge itself must first be elicited and formalized within the model. The critical distinction lies in our interactive grammatical elicitation and knowledge consolidation stages (Stage 2 \& 3), which precede scaled generation and are essential for linguistic consistency in the absence of any pre-existing digital text.

\paragraph{NLP for Low-Resource Languages} The challenge of data scarcity has led to numerous approaches, including transfer learning from high-resource languages using multilingual models such as mBERT \cite{DevlinEtAl2019} and XLM-R \cite{ConneauEtAl2020}, and dedicated multilingual translation models like NLLB \cite{CostaJussaEtAl2022}. This strategy has been applied successfully to a wide variety of languages, including Uzbek \cite{BobojonovaEtAl2025}, Urdu \cite{AminEtAl2025}, Myanmar \cite{AungDras2025}, Basque \cite{GeteEtAl2025}, Slovak \cite{ReichelEtAl2025}, Turkish \cite{YazarEtAl2025}, and Sanskrit \cite{HellwigBiagetti2025}. Closer to our specific context, similar advancements are evident in regional Northeast Indian languages. While Bodo \cite{PathakEtAl2025Bodo} and Khasi \cite{Tham2020} have seen recent progress. The lexifier language for Nagamese, Assamese has witnessed a broader development of resources, including datasets for POS tagging \cite{TalukdarSarma2024}, Named Entity Recognition \cite{PathakEtAl2022AsNER}, and conversational systems \cite{SarmaPathak2023, TamangBora2024}. Community-driven initiatives, such as Masakhane for African languages \cite{NekotoEtAl2020}, have underscored the importance of participatory research and creating resources from the ground up, a philosophy that deeply informs our work. While these methods are powerful, they often depend on some preexisting digital text, a luxury unavailable for truly ``zero-resource'' or ``low-resource'' languages like Nagamese.

\paragraph{Synthetic Data Generation} To overcome the lack of data, researchers have increasingly turned to synthetic data generation. Early methods focused on back-translation for machine translation \cite{SennrichEtAl2016}. More recently, LLMs have been used to generate labeled data from scratch \cite{SchickSchuetze2021} or to augment existing datasets for specific tasks like code-mixed translation \cite{KartikEtAl2024} and cross-lingual Named Entity Recognition \cite{LancherosEtAl2025}, with new frameworks emerging to systematically control the granularity and quality of synthetic generation \cite{SaxenaEtAl2025}. However, these approaches typically target languages or tasks that already have some existing resources. Our work distinguishes itself by proposing a complete bootstrapping pipeline in which an LLM, guided by human expertise, creates the \textit{first} foundational dataset for a language, with a strong emphasis on a human-in-the-loop validation process to ensure linguistic authenticity and mitigate model-induced artifacts.

\paragraph{NLP for Creole and Code-Switched Languages} Creole and code-switched languages present unique challenges due to their mixed linguistic origins, frequent code-switching, and lack of standardized orthography \cite{WinataEtAl2023}. While significant research has focused on widely spoken code-switched pairs like Hinglish and Spanglish \cite{GuptaEtAl2021, AguilarEtAl2020}, creole languages remain largely understudied, though recent efforts such as on Nigerian Pidgin\cite{ScholmanEtAl2025} are beginning to address this disparity. For Nagamese specifically, the only prior computational work is a CRF-based POS tagger \cite{ShoheEtAl2025}, which serves as a crucial baseline for our study. Our work provides the first large-scale, publicly available resources tailored to the specific linguistic characteristics of a creole language.

\paragraph{Fine-Tuning LLMs for Low-Resource Languages}
The current paradigm in NLP has shifted toward adapting large pre-trained models to specific domains or languages. Parameter-Efficient Fine-Tuning (PEFT) techniques, especially Low-Rank Adaptation (LoRA) \cite{HuEtAl2022}, have made it feasible to tune massive models on custom data with limited computational resources \cite{SuEtAl2024}. Several studies have explored adapting LLMs like Llama to new languages such as Persian \cite{MahdizadehSaniEtAl2025, AbbasiEtAl2023}, often highlighting challenges related to tokenization, though recent findings suggest that fine-tuning remains surprisingly effective for low-resource translation \cite{ScalviniEtAl2025}. Our work contributes to this area by providing strong empirical evidence that a model like Llama 3.2 can be effectively instruction-tuned for a low-resource creole using a primarily synthetic, human-validated dataset, demonstrating the downstream utility of our bootstrapping methodology for state-of-the-art generative tasks.

\section{Corpus Creation and Validation}

Addressing the "cold start" problem inherent in extremely low-resource languages like Nagamese Creole required us to go beyond traditional data collection methods. We introduce a novel \textbf{LLM-to-human bootstrapping pipeline}, a structured methodology designed to synthesize a high-quality corpus from scratch. This process transforms a state-of-the-art Large Language Model from a simple text generator into a linguistic partner for knowledge elicitation, grammar formalization, and scaled data production, all under rigorous human supervision. The pipeline leverages the LLM's learning capabilities for both initial text generation and preliminary annotation, which are then refined by native speakers to ensure authenticity and accuracy. The overall process is illustrated in Figure 1.
\begin{figure}[htbp] 
    \centering

    \includegraphics[width=0.4\linewidth]{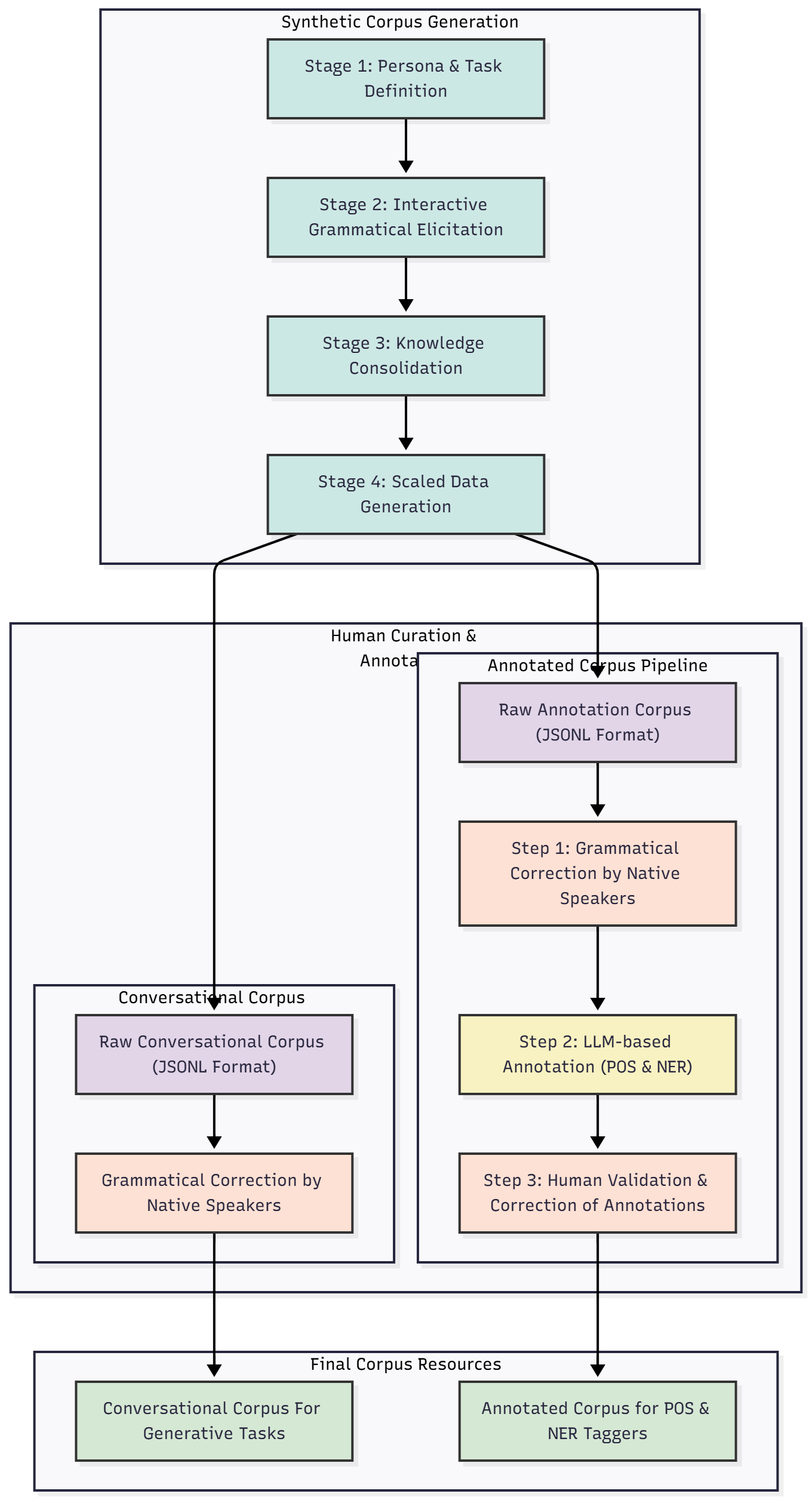} 
    
    \caption{The NagaNLP Bootstrapping Pipeline, an overview of our LLM-to-human methodology for corpus creation in a zero-resource setting.}
    \label{fig:naganlp_pipeline}
\end{figure}
\subsection{Phase 1: LLM-driven Synthetic Corpus Generation}

The foundation of our work is a synthetic corpus generated by Google's Gemini 2.5 Pro model. Rather than using simple zero-shot prompts, we employed a multi-stage conversational approach to build a robust internal representation of Nagamese within the model's context before initiating scaled generation.

\subsubsection*{Stage 1: Persona and Task Definition.}
The process began by setting a formal linguistic persona for the LLM. The initial prompt framed the task not as data generation, but as a linguistic elicitation session, positioning the model as an AI linguist tasked with learning Nagamese from a proficient speaker. This established a collaborative context for the subsequent interaction.

\subsubsection*{Stage 2: Interactive Grammatical Elicitation.}
We engaged in an iterative, interactive teaching process with the model. Rather than just providing isolated sentences, we supplied the LLM with authentic Nagamese texts, including articles and social media posts. The model's task was to process this information, ask clarifying questions about grammar and vocabulary, and form hypotheses about linguistic rules. We corrected any mistakes it made during this learning phase, providing direct feedback and corrected examples. This interactive loop allowed the model to build a progressively complex and accurate model of Nagamese syntax and morphology from first principles, mirroring techniques from field linguistics.

\subsubsection*{Stage 3: Knowledge Consolidation.}
After the elicitation phase, we prompted the model to synthesize its learned knowledge into a comprehensive, structured grammar of Nagamese. This step forced the model to consolidate its scattered conversational learnings into a coherent knowledge base, which proved crucial for maintaining linguistic consistency during scaled generation.

\subsubsection*{Stage 4: Scaled Data Generation with In-Context Reinforcement.}
With a robust understanding of Nagamese established, we shifted to scaled data production for two separate targets. We used structured, few-shot prompting strategies to generate a large set of conversational pairs (for the generative dataset) and a separate set of diverse declarative sentences (for the annotation dataset), both in a clean JSONL format. To prevent context drift and maintain quality over thousands of generations, we periodically reinforced the model’s knowledge by re-injecting core grammatical rules and authentic Nagamese texts, such as newspaper editorials, between generation batches. This process yielded the raw text for both the annotated and conversational corpora.

\subsection{Human Curation and Annotation}

The credibility of any supervised model hinges on the quality of its training data. To ensure the linguistic authenticity and accuracy of our synthetically-generated corpora, we implemented a rigorous multi-stage human-in-the-loop protocol for both curation and annotation.

\subsubsection*{Annotator Profile and Training.}
The validation and annotation tasks were performed by a team of four annotators, including the primary author. Our annotation team consisted of four Nagamese Creole speakers (three native and one fluent), providing reliable handling of idiomatic usage, dialectal nuance, and code-switching phenomena. Prior to annotation, all participants completed a training session on the annotation guidelines and tool to standardize the process.

\subsubsection*{Corpus Curation and Annotation Process.}
The two corpora were finalized through distinct pipelines:
\begin{itemize}
    \item \textbf{The NagaNLP Conversational Corpus} underwent a single, crucial human validation step. The raw generated conversational pairs were reviewed and corrected by native speakers to fix any grammatical errors, unnatural phrasing, or logical inconsistencies.
    \item \textbf{The NagaNLP Annotated Corpus} was created through a three-step process. First, like the conversational corpus, the raw generated sentences were corrected for grammatical accuracy by native speakers. Second, these cleaned sentences were passed back to the Nagamese-aware LLM, which performed an initial annotation pass for both Part-of-Speech (POS) tags and Named Entities (NER). Finally, the LLM-generated annotations were meticulously reviewed, corrected, and finalized by the human annotators to create the gold-standard corpus.
\end{itemize}

\subsubsection*{Annotation Schema.}
\begin{itemize}
    \item \textbf{Part-of-Speech (POS) Tagging:} We adopted the Universal Dependencies (UD) v2 tagset \cite{NivreEtAl2020}, a standard framework with 17 universal tags that facilitates cross-linguistic research and aligns with recent advancements in parallel syntactic representations \cite{AlzettaEtAl2025}. Code-switched English words were tagged based on their grammatical function within the Nagamese sentence (e.g., \textit{situation} as \texttt{NOUN}).
    \item \textbf{Named Entity Recognition (NER):} We used the IOB2 format \cite{TjongKimSangVeenstra1999} to annotate four standard entity types: \texttt{PER} (Person), \texttt{LOC} (Location), \texttt{ORG} (Organization), and \texttt{MISC} (Miscellaneous). The annotated corpus is a single resource; for NER tasks, both POS and NER tags are available, while for POS tagging tasks, only the POS tags are utilized.
\end{itemize}

\subsubsection*{Inter-Annotator Agreement (IAA).}
To validate our final annotation schema and ensure data reliability after the human review stage, a sample of 200 sentences was independently annotated from scratch by two trained speakers. We measured agreement using Cohen's Kappa ($\kappa$)\cite{Cohen1960}, which accounts for chance agreement. We achieved a Kappa score of \textbf{$\kappa$ = 0.92} for POS tagging and \textbf{$\kappa$ = 0.88} for NER. These scores indicate near-perfect and substantial agreement, respectively, confirming the high quality and consistency of our final human-validated annotations. The disagreements were adjudicated by an expert annotator to create the final test set.

\subsection{Corpus Statistics and Data Splits}

Our pipeline produced two distinct corpora: the \textbf{NagaNLP Annotated Corpus} for POS/NER tasks and the \textbf{NagaNLP Conversational Corpus} for LLM fine-tuning. Both were partitioned into 80\% training, 10\% development, and 10\% held-out test sets. The final statistics for the corpora are detailed in Table 1. The NagaNLP Annotated Corpus consists of 214 sentences (4,839 tokens) densely annotated for both POS and NER. The distribution of POS tags (Table 2) reveals linguistic characteristics of Nagamese, such as the frequent use of the possessive marker \textit{laga}, tagged as \texttt{PART}. The NER distribution (Table 3) shows a focus on \texttt{MISC} and \texttt{LOC} entities, reflecting the source domains of the initial generation seeds. The NagaNLP Conversational Corpus is significantly larger, containing 10,018 instruction pairs and more than 300,000 tokens, providing a substantial resource for training generative models.

\begin{table}[htbp]
\centering
\caption{Overall statistics for the \texttt{NagaNLP} corpora after splitting.}
\begin{tabular}{l c c c}
\hline
\textbf{Metric} & \textbf{POS Corpus} & \textbf{NER Corpus} & \textbf{LLM Corpus} \\
\hline
\textbf{Total Sentences/Pairs} & \textbf{214} & \textbf{214} & \textbf{10,018} \\
\hline
Train/Dev/Test Split & 171/21/22 & 171/21/22 & 8,014/1,002/1,002 \\
\hline
\textbf{Total Tokens} & \textbf{4,839} & \textbf{4,839} & \textbf{311,684} \\
\hline
Vocabulary Size & 1,515 & 1,515 & 22,998 \\
\hline
\end{tabular}
\label{tab:corpus_stats}
\end{table}

\begin{table}[htbp]
\centering
\caption{Part-of-Speech (POS) tag distribution in the annotated corpus based on universal dependencies.}
\begin{tabular}{l r r}
\toprule
\textbf{Tag} & \textbf{Count} & \textbf{Percentage (\%)} \\
\toprule
NOUN & 921 & 19.03\% \\
VERB & 792 & 16.37\% \\
PROPN & 711 & 14.69\% \\
PUNCT & 582 & 12.03\% \\
ADP & 425 & 8.78\% \\
PART & 384 & 7.94\% \\
ADJ & 314 & 6.49\% \\
NUM & 204 & 4.22\% \\
CCONJ & 168 & 3.47\% \\
PRON & 164 & 3.39\% \\
ADV & 132 & 2.73\% \\
SCONJ & 41 & 0.85\% \\
DET & 1 & 0.02\% \\
\bottomrule
\end{tabular}
\label{tab:pos_dist}
\end{table}

\begin{table}[htbp]
\centering
\caption{Named Entity Recognition (NER) entity distribution (IOB2 format)}
\begin{tabular}{l r r}
\toprule
\textbf{Tag} & \textbf{Count} & \textbf{Percentage (\%)} \\
\midrule
MISC & 172 & 36.36\% \\
LOC & 151 & 31.92\% \\
PER & 77 & 16.28\% \\
ORG & 73 & 15.43\% \\
\bottomrule
\end{tabular}
\label{tab:ner_dist}
\end{table}

\section{Experimental Setup}
To empirically validate the quality of our annotated resources and the efficacy of our pipeline, we designed a comprehensive experimental framework addressing both foundational and generative capabilities. We conducted experiments across two distinct categories: (1) \textbf{Foundational Tasks}, specifically Part-of-Speech (POS) tagging and Named Entity Recognition (NER), to benchmark the \textit{NagaNLP Annotated Corpus}; and (2) \textbf{Generative Tasks}, utilizing the \textit{NagaNLP Conversational Corpus} to fine-tune a state-of-the-art Large Language Model (LLM) for instruction following, summarization, and translation. All experiments were rigorously evaluated using the data splits detailed in Table \ref{tab:corpus_stats} against strong statistical and neural baselines.

\subsection{Foundational Tasks: POS Tagging and NER}

This section details the experimental setup for fine-tuning foundational models on our annotated Nagamese corpus for Part-of-Speech (POS) tagging and Named Entity Recognition (NER). It outlines the model architectures, hyperparameters, baseline comparisons, and final results that validate the quality of the corpus.

\subsubsection{Model Architecture and Hyperparameters}
We fine-tuned two widely used pre-trained multilingual transformer models for the token classification tasks: \texttt{bert-base-multilingual-cased} and \texttt{xlm-roberta-base}. For our final reported results, we selected the best-performing architecture for each respective task based on the macro F1-score achieved on the development set, following recent work demonstrating the efficacy of transformer models for POS tagging \cite{LiEtAl2022}. All transformer models were fine-tuned using the following hyperparameters: \textbf{Optimizer:} AdamW\cite{LoshchilovHutter2019}, \textbf{Learning Rate:} 2e-5, \textbf{Weight Decay:} 0.01, \textbf{Batch Size:} 16, \textbf{Training Epochs:} 20. We implemented an epoch-based evaluation strategy, saving only the model checkpoint that achieved the highest macro F1-score on the development set.

\subsubsection{Baselines}
We benchmark our models against a robust set of baselines to contextualize their performance:
\begin{enumerate}
    \item \textbf{Zero-Shot XLM-R:} We evaluate the zero-shot performance of \texttt{xlm-roberta-large\cite{ConneauEtAl2020}} on our held-out test set. This establishes a "zero-resource" baseline, measuring the model’s ability to transfer its knowledge to Nagamese without any task-specific fine-tuning.
    \item \textbf{CRF (Prior Work):} We report the accuracy and F1-score from the only known prior work on Nagamese POS tagging, which employed a Conditional Random Fields (CRF) model \cite{ShoheEtAl2025}.
    \item \textbf{CRF (Our Data):} To provide a direct and fair comparison against a strong statistical method, we replicated the feature engineering of prior work and trained a CRF model on our own training data split. This baseline effectively controls for the dataset, isolating the performance contribution of the transformer architecture versus the quality of the corpus itself.
\end{enumerate}

\subsubsection{Evaluation Metrics}
\textbf{For POS tagging,} we report overall accuracy and macro-averaged Precision, Recall, and F1-score. \textbf{For NER,} we report the standard entity-level strict Accuracy, Precision, Recall, and F1-score using the \texttt{seqeval} framework (IOB2 scheme)\cite{seqeval}, which correctly evaluates chunk-based annotations.

\subsection{Generative Task: Instruction Fine-Tuning}

To demonstrate the utility of our NagaNLP Conversational Corpus for state-of-the-art generative tasks, we fine-tuned a powerful instruction-based Large Language Model and conducted a comprehensive evaluation covering conversational ability, summarization, and machine translation.

\subsubsection{Model and Fine-Tuning.} Our experiments are centered around \textbf{NagaLLaMA}, our fine-tuned version of \texttt{meta-llama/Llama-3.2-3B-Instruct}. We employed Parameter-Efficient Fine-Tuning (PEFT) using Low-Rank Adaptation (LoRA) \cite{HuEtAl2022} to make training computationally tractable. The LoRA configuration was set with a rank (\textit{r}) of 16, an alpha of 32, and a dropout rate of 0.05. The adaptation was applied to a comprehensive set of target modules within the transformer architecture: \texttt{q\_proj}, \texttt{k\_proj}, \texttt{v\_proj}, \texttt{o\_proj}, \texttt{gate\_proj}, \texttt{up\_proj}, and \texttt{down\_proj}. The model was trained for 3 epochs using a learning rate of 2e-4, a per-device batch size of 2, and gradient accumulation steps of 8, resulting in an effective batch size of 16.

\subsubsection{Baselines.} The performance of NagaLLaMA is benchmarked against two strong baselines:
\begin{enumerate}
    \item \textbf{Llama 3.2-3B (Few-Shot):} The base \texttt{Llama-3.2-3B-Instruct} model was evaluated on our test set using a 3-shot in-context learning prompt. This baseline measures the pre-trained model's ability to perform Nagamese tasks without any weight updates.
    \item \textbf{NLLB-200 (Translation):} For the specific task of English-to-Nagamese translation, we use the \texttt{facebook/nllb-200-distilled-600M} model \cite{CostaJussaEtAl2022}. As NLLB does not support Nagamese, we use Assamese (\texttt{asm\_Beng}) as the target language, a common proxy for Nagamese in multilingual models.
\end{enumerate}

\subsubsection{Evaluation Metrics.} We employ a suite of automatic metrics to evaluate performance across different tasks:
\begin{itemize}
    \item \textbf{Perplexity (PPL)\cite{BrownEtAl1990}:} To measure the model's overall linguistic fluency and predictive accuracy on the held-out test set. A lower score is better.
    \item \textbf{ROUGE-L\cite{Lin2004ROUGE}:} To evaluate performance in summarization and general conversational tasks by measuring the longest common subsequence between generated text and references.
    \item \textbf{BLEU \& chrF++\cite{PapineniEtAl2002BLEU, Popovic2017chrFpp}:} Standard metrics for evaluating machine translation quality, measuring n-gram precision and character n-gram F-score, respectively.
    \item \textbf{COMET\cite{ReiEtAl2020}:} A state-of-the-art neural metric that evaluates translation quality by measuring the semantic similarity between the source, machine translation and the reference.
\end{itemize}

\section{Results and Analysis}

Our experiments confirm the efficacy of our data generation and annotation pipeline. The models trained on the NagaNLP corpus establish a new state-of-the-art for Nagamese, significantly outperforming all baseline models.

\subsection{Part-of-Speech Tagging Results}

The models trained on our annotated data demonstrate a profound understanding of Nagamese grammar. As shown in Table~\ref{tab:pos_main_results}, both fine-tuned transformers set a new SOTA, while the performance of the CRF baseline trained on our data highlights the quality of the corpus itself.

\begin{table}[ht!]
\centering
\caption{Main results for the Part-of-Speech (POS) Tagging task. Best performance is in \textbf{bold}.}
\label{tab:pos_main_results}
\begin{tabular}{@{}lrr@{}}
\toprule
\textbf{Model} & \textbf{Accuracy} & \textbf{F1-Score (Macro)} \\ \midrule
Zero-Shot XLM-R (\texttt{large}) & 14.69\% & 0.02 \\
CRF (Shohe et al., 2025) & 85.70\% & 0.86 \\
CRF (Our Data) & 93.84\% & 0.91 \\ \midrule
\textit{NagaNLP Transformers (Ours)} \\
\quad \texttt{xlm-roberta-base} & 92.92\% & 0.88 \\
\quad \texttt{bert-base-multilingual-cased} & \textbf{93.81\%} & \textbf{0.90} \\ \bottomrule
\end{tabular}
\end{table}

The \texttt{bert-base-multilingual-cased} model emerged as the top performer, achieving a final accuracy of \textbf{93.81\%} and a macro F1-score of \textbf{0.90}. This result surpasses both the alternative \texttt{xlm-roberta-base} transformer (0.88 F1) and the previous benchmark set by Shohe et al. (2025). The zero-shot baseline performs at a near-random chance level (0.02 F1), confirming that large multilingual models possess no inherent knowledge of Nagamese syntax and validating the necessity of our corpus.

\begin{figure}[htbp]
    \centering % Center the entire figure content

    % --- Top Row of Images ---
    \begin{subfigure}{0.48\textwidth}
        \includegraphics[width=\linewidth]{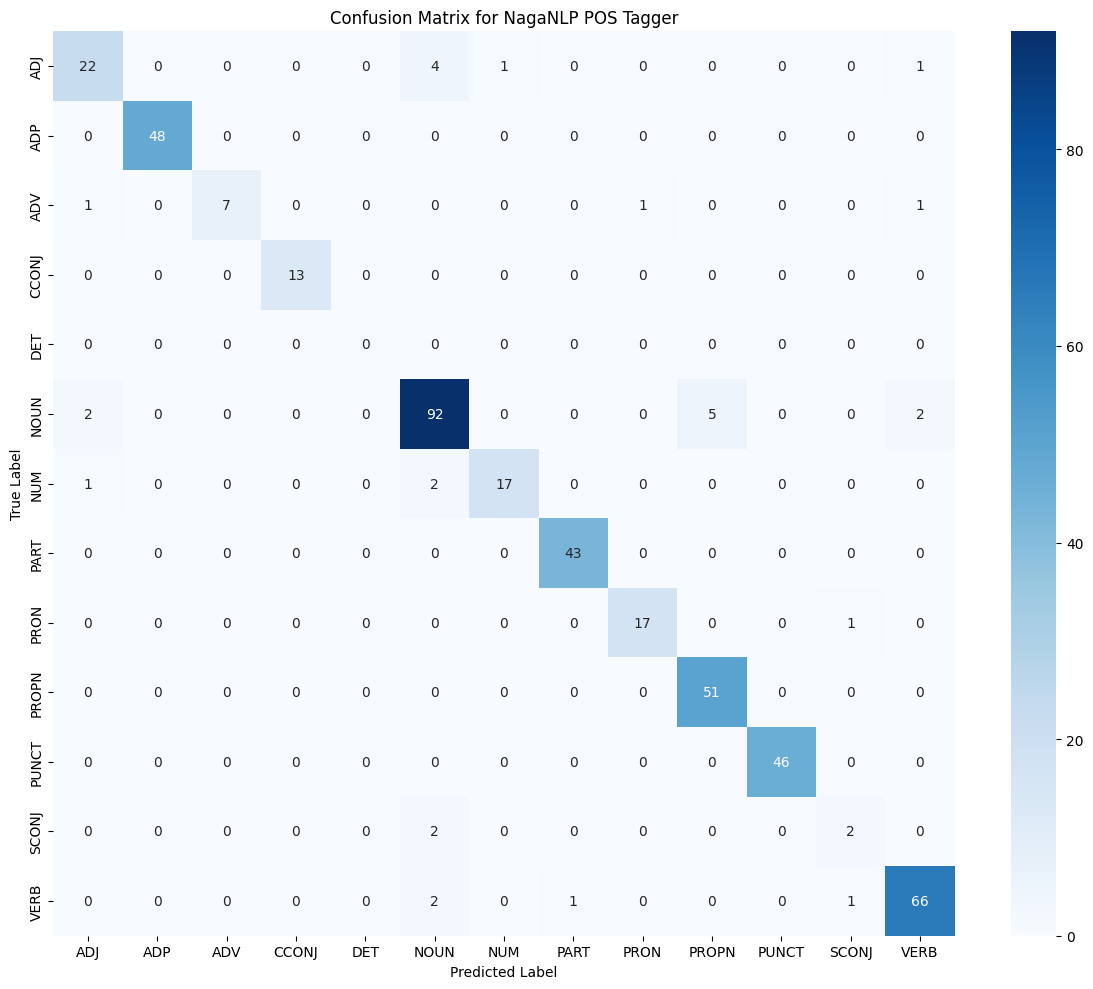}
        \subcaption{NagaNLP Transformer (\texttt{bert-base-multilingual-cased})}
        \label{subfig:bert_matrix}
    \end{subfigure}
    \hfill % This creates a horizontal space between the two images
    \begin{subfigure}{0.48\textwidth}
        \includegraphics[width=\linewidth]{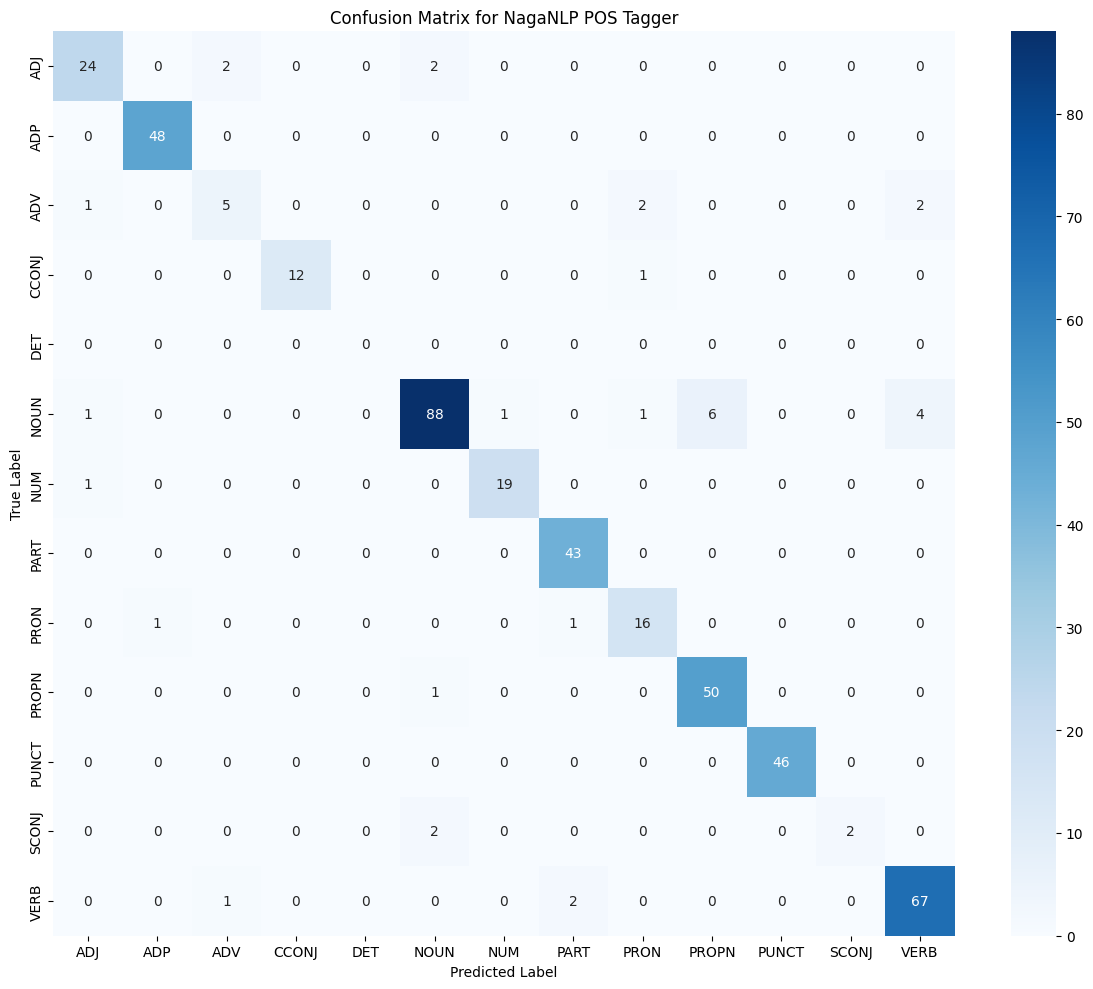}
        \subcaption{NagaNLP Transformer (\texttt{xlm-roberta-base})}
        \label{subfig:xlmr_matrix}
    \end{subfigure}

    \vspace{1em} % Adds a little vertical space between the rows

    % --- Bottom Row of Images ---
    \begin{subfigure}{0.48\textwidth}
        \includegraphics[width=\linewidth]{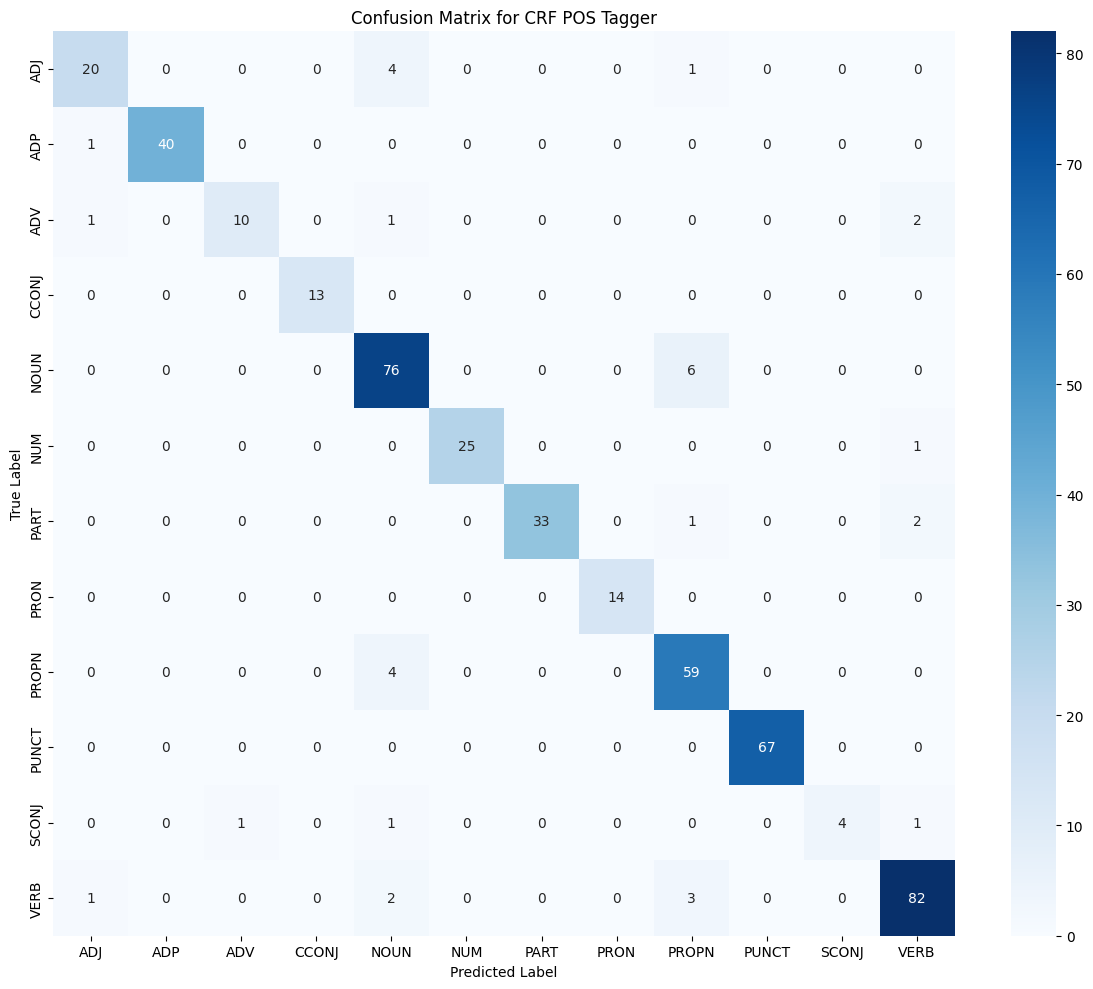}
        \subcaption{CRF Baseline (Our Data)}
        \label{subfig:crf_matrix}
    \end{subfigure}
    \hfill % Horizontal space
    \begin{subfigure}{0.48\textwidth}
        \includegraphics[width=\linewidth]{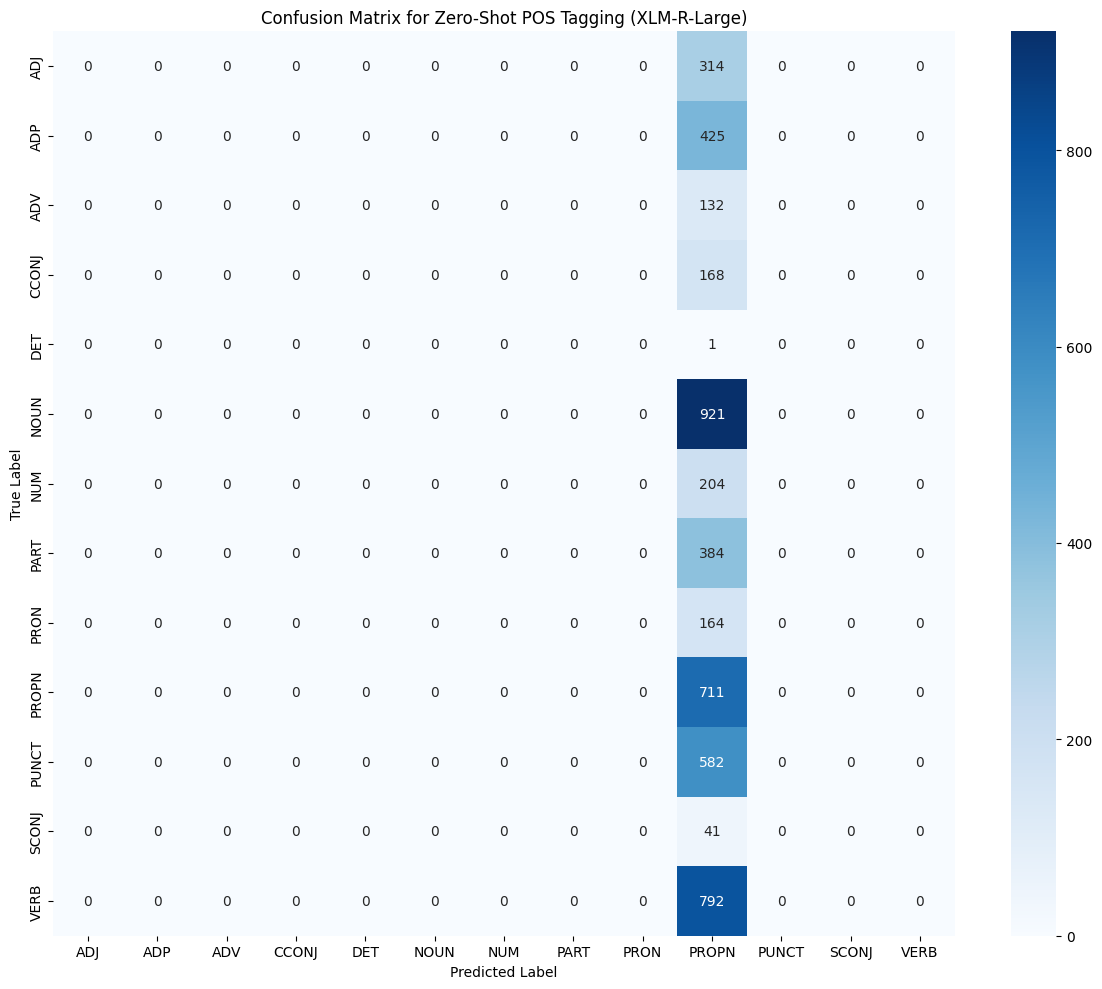}
        \subcaption{Zero-Shot Baseline (\texttt{xlm-roberta-large})}
        \label{subfig:zeroshot_matrix}
    \end{subfigure}

    \caption{Comparison of confusion matrices for the Part-of-Speech (POS) tagging task across all evaluated models. The fine-tuned transformer models (a, b) and the CRF baseline trained on our data (c) show strong diagonal alignment, indicating high accuracy. In contrast, the zero-shot baseline (d) fails to correctly classify any tags, confirming its lack of inherent knowledge of Nagamese syntax.}
    \label{fig:pos_confusion_matrices_all}
\end{figure}

A crucial finding is the performance of the CRF model trained on our data, which achieves a \textbf{0.91 F1-score}. The fact that a traditional statistical model performs on par with a fine-tuned transformer is a strong testament to the high quality, consistency, and linguistic richness of our human-validated corpus. The detailed per-tag performance of our best transformer model (\texttt{bert-base-multilingual-cased}) is shown in Table~\ref{tab:pos_per_tag_report}.

\begin{table}[ht!]
\centering
\caption{Per-tag classification report for the NagaNLP POS Tagger (\texttt{bert-base-multilingual-cased}).}
\label{tab:pos_per_tag_report}
\begin{tabular}{@{}lrrrr@{}}
\toprule
\textbf{Tag} & \textbf{Precision} & \textbf{Recall} & \textbf{F1-Score} & \textbf{Support} \\ \midrule
ADJ & 0.85 & 0.79 & 0.81 & 28 \\
ADP & 1.00 & 1.00 & 1.00 & 48 \\
ADV & 1.00 & 0.70 & 0.82 & 10 \\
CCONJ & 1.00 & 1.00 & 1.00 & 13 \\
NOUN & 0.90 & 0.91 & 0.91 & 101 \\
NUM & 0.94 & 0.85 & 0.89 & 20 \\
PART & 0.98 & 1.00 & 0.99 & 43 \\
PRON & 0.94 & 0.94 & 0.94 & 18 \\
PROPN & 0.91 & 1.00 & 0.95 & 51 \\
PUNCT & 1.00 & 1.00 & 1.00 & 46 \\
SCONJ & 0.50 & 0.50 & 0.50 & 4 \\
VERB & 0.94 & 0.94 & 0.94 & 70 \\ \bottomrule
\end{tabular}
\end{table}

The model shows robust performance across most categories, with perfect or near-perfect scores for function tags like \texttt{ADP}, \texttt{CCONJ}, \texttt{PART}, and \texttt{PUNCT}. The lowest performance is on \texttt{SCONJ} (Subordinating Conjunction), which is expected given its very low support (only 4 instances) in the test set.

\subsection{Named Entity Recognition Results}

For NER, our models establish the first-ever benchmarks for Nagamese. In this task, the \texttt{xlm-roberta-base} model proved most effective, significantly outperforming \texttt{bert-base-multilingual-cased}, as detailed in Table~\ref{tab:ner_main_results}.

\begin{table}[ht!]
\centering
\caption{Main results for the Named Entity Recognition (NER) task. Best performance is in \textbf{bold}.}
\label{tab:ner_main_results}
\begin{tabular}{@{}lrr@{}}
\toprule
\textbf{Model} & \textbf{Accuracy (Strict)} & \textbf{F1-Score (Macro)} \\ \midrule
Zero-Shot XLM-R (\texttt{large}) & 0.00\% & $\sim$0.00 \\ \midrule
\textit{NagaNLP Transformers (Ours)} \\
\quad \texttt{bert-base-multilingual-cased} & 95.13\% & 0.57 \\
\quad \texttt{xlm-roberta-base} & \textbf{95.13\%} & \textbf{0.75} \\ \bottomrule
\end{tabular}
\end{table}

Our best model (\texttt{xlm-roberta-base}) achieves a strict, entity-level accuracy of \textbf{95.13\%} and a macro F1-score of \textbf{0.75}, a strong result for a task being defined for the first time in this language. The \texttt{bert-base-multilingual-cased} model reached the same accuracy but struggled with precision and recall, yielding a much lower F1-score of 0.57. The zero-shot baseline fails completely, scoring a macro F1 of essentially zero, which underscores that the model's NER capability was learned exclusively from our dataset. The detailed per-entity breakdown in Table~\ref{tab:ner_per_entity_report} shows the ability of our best model to distinguish between different entity types.

\begin{table}[ht!]
\centering
\caption{Per-entity classification report for the NagaNLP NER model (\texttt{xlm-roberta-base}).}
\label{tab:ner_per_entity_report}
\begin{tabular}{@{}lrrrr@{}}
\toprule
\textbf{Entity} & \textbf{Precision} & \textbf{Recall} & \textbf{F1-Score} & \textbf{Support} \\ \midrule
LOC & 0.80 & 1.00 & 0.89 & 4 \\
MISC & 0.55 & 0.85 & 0.67 & 13 \\
ORG & 0.50 & 0.57 & 0.53 & 7 \\
PER & 1.00 & 0.86 & 0.92 & 7 \\ \bottomrule
\end{tabular}
\end{table}

The model performs exceptionally well on \texttt{PER} (Person, 0.92 F1) and \texttt{LOC} (Location, 0.89 F1), which are typically well-defined and syntactically distinct. Performance is lower for \texttt{ORG} (Organization, 0.53 F1) and the more semantically diverse \texttt{MISC} (Miscellaneous, 0.67 F1). This is a common challenge in NER tasks, likely exacerbated by the limited number of examples for these classes in our initial corpus. These results collectively validate our data creation pipeline as a highly effective method for building foundational NLP resources in a zero-resource setting.

\subsection{Generative Model Performance}

Our fine-tuned model, NagaLLaMA, demonstrates a transformative improvement in its ability to understand and generate Nagamese compared to the base model, validating the high quality and effectiveness of our conversational corpus.

\paragraph{Overall Performance.} As shown in Table~\ref{tab:generative_results}, NagaLLaMA achieves a perplexity of \textbf{3.85} on the test set, a dramatic reduction from the few-shot baseline's score of 96.76. This indicates a profound improvement in the model's fundamental grasp of Nagamese syntax, semantics, and conversational patterns. Similarly, its ROUGE-L score of \textbf{20.77} nearly doubles that of the baseline, showing a significantly enhanced ability to generate relevant and coherent responses.

\begin{table}[ht!]
\centering
\caption{Automatic evaluation results for generative models. NagaLLaMA significantly outperforms the base model on core metrics.}
\label{tab:generative_results}
\begin{tabular}{@{}lrr@{}}
\toprule
\textbf{Model} & \textbf{Perplexity (PPL) $\downarrow$} & \textbf{ROUGE-L $\uparrow$} \\ \midrule
Llama 3.2-3B (Few-Shot) & 96.76 & 11.28 \\
\textbf{NagaLLaMA (Ours)} & \textbf{3.85} & \textbf{20.77} \\ \bottomrule
\end{tabular}
\end{table}

\paragraph{Machine Translation Performance.} We conducted a detailed evaluation on a dedicated test set of 259 English-Nagamese parallel sentences. The results, presented in Table~\ref{tab:mt_results}, highlight the superiority of our specialized model.

\begin{table}[ht!]
\centering
\caption{Automatic evaluation results for the machine translation task (English $\leftrightarrow$ Nagamese).}
\label{tab:mt_results}
\begin{tabular}{@{}lrrr@{}}
\toprule
\textbf{Model \& Direction} & \textbf{BLEU $\uparrow$} & \textbf{chrF++ $\uparrow$} & \textbf{COMET $\uparrow$} \\ \midrule
NLLB-200 (Eng $\rightarrow$ Nag) & 1.64 & 0.30 & 0.5875 \\
NLLB-200 (Nag $\rightarrow$ Eng) & 2.23 & 18.71 & 0.4227 \\
\textbf{NagaLLaMA (Eng $\rightarrow$ Nag)} & \textbf{14.25} & \textbf{41.83} & \textbf{0.6668} \\
\textbf{NagaLLaMA (Nag $\rightarrow$ Eng)} & \textbf{34.97} & \textbf{53.17} & \textbf{0.7338} \\ \bottomrule
\end{tabular}
\end{table}

The NLLB-200 model, using Assamese as a proxy, does not produce meaningful translations into Nagamese, achieving a near-zero BLEU score of 1.64. In contrast, \textbf{NagaLLaMA} demonstrates competent translation capabilities, achieving a BLEU score of \textbf{14.25} for English-to-Nagamese and a very strong \textbf{34.97} for Nagamese-to-English. The high chrF++ and COMET scores further confirm that NagaLLaMA generates translations that are not only lexically similar but also semantically coherent. This stark difference proves that fine-tuning on our targeted, high-quality synthetic data is vastly superior to relying on proxy languages in large multilingual models for this low-resource creole.

\subsection{Ablation Studies}
\label{sec:ablation}

To better understand the key components of our methodology and validate our design choices, we conduct two critical ablation studies. The first investigates the impact of our human-in-the-loop validation process on foundational task performance, while the second analyzes the effect of data scale on the generative model.

\subsubsection{Impact of Human-in-the-Loop (HiTL) Validation}
\label{ssec:hitl_ablation}

A core claim of our work is that raw, LLM-generated synthetic data, while a valuable starting point, is insufficient for building high-quality models without rigorous human oversight. To quantify the contribution of our human validation and correction phase, we perform an ablation on the foundational POS and NER tasks.

\textbf{Setup.} We compare our main \texttt{NagaNLP Transformer} models against identical models trained on a ``Raw Synthetic'' version of the corpus. This dataset consists of the initial text generated by Gemini 2.5 and annotated using a zero-shot LLM prompt, but \textit{before} any review, correction, or re-annotation by our native-speaking annotators.

\textbf{Results.} The results, presented in Table~\ref{tab:hitl_ablation}, show a substantial performance degradation when the human-in-the-loop component is removed.

\begin{table}[ht!]
\centering
\caption{Ablation on Human-in-the-Loop (HiTL) Validation. Performance of models trained on raw synthetic data vs. our final human-validated corpus. Scores are F1-Macro.}
\label{tab:hitl_ablation}
\begin{tabular}{@{}lcc@{}}
\toprule
\textbf{Model \& Training Data}            & \textbf{POS (F1-Macro)} & \textbf{NER (F1-Macro)} \\ \midrule
Transformer (Raw Synthetic Data)         & 0.81                    & 0.62                    \\
\textbf{Transformer (Human-Validated)} & \textbf{0.90}           & \textbf{0.75}           \\ \bottomrule
\end{tabular}
\end{table}

On POS tagging, the model trained on the final, validated corpus outperforms the one trained on raw data by 9 F1 points. The gap is even more pronounced for NER, with a 13-point F1 improvement. This confirms our hypothesis that the HiTL stage is critical. Qualitative analysis of the raw data revealed common errors such as unnatural phrasing, subtle grammatical mistakes, and incorrect entity boundary predictions, all of which were rectified by our human annotators. These results provide strong quantitative evidence for the necessity of human oversight in synthetic data generation pipelines for low-resource languages.

\subsubsection{Effect of Data Scale on Generative Performance}
\label{ssec:scaling_ablation}

Our methodology was designed to be scalable, culminating in a 10K-pair conversational corpus. To analyze the relationship between the volume of this synthetic-hybrid data and the performance of NagaLLaMA, we trained the model on incremental fractions of the training set.

\textbf{Setup.} We fine-tuned NagaLLaMA on 25\%, 50\%, 75\%, and 100\% of the 8,014-pair training set and measured the resulting perplexity and evaluation loss on the held-out validation set.

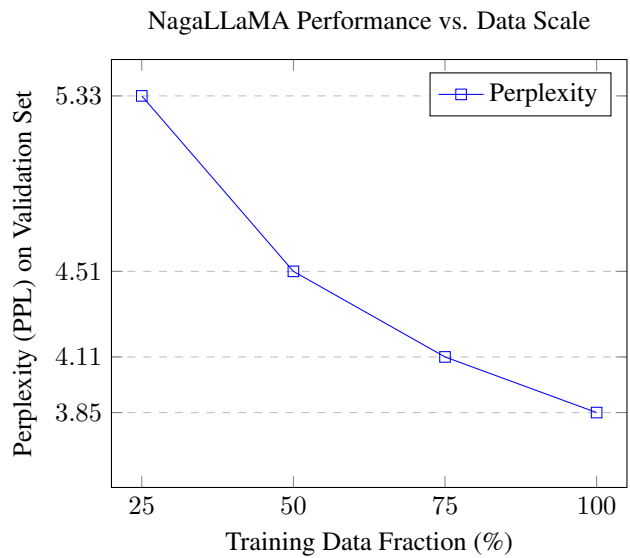
\begin{figure}[htbp]
\centering
\begin{tikzpicture}
\begin{axis}[
    title={NagaLLaMA Performance vs. Data Scale},
    xlabel={Training Data Fraction (\%)},
    ylabel={Perplexity (PPL) on Validation Set},
    xmin=20, xmax=105,
    ymin=3.5, ymax=5.5,
    xtick={25,50,75,100},
    ytick={3.85, 4.11, 4.51, 5.33},
    legend pos=north east,
    ymajorgrids=true,
    grid style=dashed,
]
\addplot[
    color=blue,
    mark=square,
    ]
    coordinates {
    (25,5.33)(50,4.51)(75,4.11)(100,3.85)
    };
    \legend{Perplexity}
\end{axis}
\end{tikzpicture}
\caption{Data scaling curve for NagaLLaMA. Perplexity consistently decreases as the volume of training data increases, highlighting the value of our full conversational corpus.}
\label{fig:scaling_curve}
\end{figure}

\textbf{Results.} As detailed in Table~\ref{tab:data_scaling_ablation} and visualized in Figure~\ref{fig:scaling_curve}, there is a clear and consistent trend: model performance improves steadily as more training data is used.

\begin{table}[ht!]
\centering
\caption{Ablation study on the impact of training data scale on NagaLLaMA's performance. Perplexity and loss decrease consistently as more of our synthetic-hybrid training data is used. Metrics are reported on the validation set.}
\label{tab:data_scaling_ablation}
\begin{tabular}{@{}cccc@{}}
\toprule
\textbf{Data Fraction (\%)} & \textbf{Train Samples} & \textbf{Perplexity (PPL) $\downarrow$} & \textbf{Eval Loss $\downarrow$} \\ \midrule
25                        & 2,004                  & 5.33                                  & 1.67                          \\
50                        & 4,007                  & 4.51                                  & 1.51                          \\
75                        & 6,011                  & 4.11                                  & 1.41                          \\
100                       & 8,014                  & 3.85                                  & 1.35                          \\ \bottomrule
\end{tabular}
\end{table}

The model's perplexity drops from 5.33 when trained on just 25\% of the data to 3.85 when using the full corpus. This learning curve demonstrates that the model's fluency and predictive grasp of Nagamese are strongly correlated with the amount of our high-quality training data it is exposed to. The consistent improvement across all fractions validates the effectiveness of generating a larger-scale (10K pair) dataset. Furthermore, the fact that performance has not yet plateaued suggests that generating even more data with our pipeline could lead to further gains.

\section{Conclusion}

This paper confronted the critical data scarcity problem for the low-resource Nagamese Creole by introducing a novel and efficient LLM-to-human bootstrapping pipeline. Our methodology successfully leverages a state-of-the-art LLM as a "language elicitor" under expert guidance, coupled with a rigorous human-in-the-loop validation process, to generate high-quality annotated and conversational corpora from a zero-resource starting point. The empirical validation of our methodology is comprehensive and unequivocal. We used the generated data to build NagaNLP, the first open-source NLP toolkit for Nagamese. Our foundational models set a new state-of-the-art, achieving a 0.90 macro F1 score for part-of-speech tagging and establishing the first benchmark with a 0.75 macro F1 score for Named Entity Recognition. Furthermore, we demonstrated the downstream utility of our conversational data by fine-tuning a 3B parameter LLM, NagaLLaMA, which achieved a perplexity of 3.85, an order of magnitude improvement over its few-shot counterpart, and showed strong performance on machine translation, far surpassing dedicated multilingual models. These results confirm our central hypothesis: a human-validated, synthetic-hybrid corpus can effectively bootstrap a full suite of modern NLP tools for a previously undigitized language. In releasing the NagaNLP toolkit, including all datasets, models, and code, we provide not only the first foundational resources for Nagamese, but also a replicable blueprint for researchers and communities working to bridge the digital divide for other underresourced languages.

%Bibliography
\bibliographystyle{unsrt}  
\bibliography{references}

\begin{thebibliography}{10}

\bibitem{JoshiEtAl2020}
Pratik Joshi, Sebastin Santy, Amar Budhiraja, Kalika Bali, and Monojit Choudhury.
\newblock The state and fate of linguistic diversity and inclusion in the {NLP} world.
\newblock In {\em Proceedings of the 58th Annual Meeting of the Association for Computational Linguistics}, pages 6282--6293, Online, 2020. Association for Computational Linguistics.

\bibitem{Kornai2013}
Andras Kornai.
\newblock Digital language death.
\newblock {\em PLOS ONE}, 8(10):1--11, 2013.

\bibitem{jayan2025challenges}
Jisha~P Jayan, J~Satheesh Kumar, and T~Amudha.
\newblock Challenges and improvisation in machine translation: the case of malayalam--tamil machine translation.
\newblock {\em Language Resources and Evaluation}, pages 1--34, 2025.

\bibitem{shafer2025building}
Omri Shafer-Raviv, Or~Aleksandrowicz, Nick Howell, and Daniel Rosenberg.
\newblock Building a specialised hebrew textual corpus on construction, planning and architecture.
\newblock {\em Language Resources and Evaluation}, pages 1--21, 2025.

\bibitem{BaruahSarma2023}
Niyor Baruah and Shikhar~Kumar Sarma.
\newblock Parts-of-speech tagger in {Assamese} using {LSTM} and {Bi-LSTM}.
\newblock In {\em Proceedings of the International Conference on Advances in Data-driven Computing and Intelligent Systems}, pages 535--544. Springer, 2023.

\bibitem{DasSingh2024}
Ringki Das and Thoudam~Doren Singh.
\newblock Which words are important?: An empirical study of {Assamese} sentiment analysis.
\newblock {\em Language Resources and Evaluation}, 58:1--24, 2024.

\bibitem{BaruahHazarika2014}
Udayan Baruah and Shyamanta~M. Hazarika.
\newblock A dataset of online handwritten {Assamese} characters.
\newblock {\em Journal of Information Processing Systems}, 10(4):631--649, 2014.

\bibitem{ChoudhuryEtAl2024}
Pankaj Choudhury, Prithwijit Guha, and Sukumar Nandi.
\newblock Impact of language-specific training on image caption synthesis: A case study on low-resource {Assamese} language.
\newblock {\em International Journal of Asian Language Processing}, 34(01):2450002, 2024.

\bibitem{ShoheEtAl2025}
A.~N. Shohe, C.~Khiamungam, and T.~Angami.
\newblock Part-of-speech tagging for {Nagamese} language using {CRF}, 2025.

\bibitem{MagueresseEtAl2020}
Alexandre Magueresse, Vincent Carles, and Evan Heetderks.
\newblock Low-resource languages: A review of past trends and future challenges.
\newblock {\em arXiv preprint arXiv:2006.07264}, 2020.

\bibitem{VerdonikEtAl2025}
Darinka Verdonik, Andreja Bizjak, Andrej {\v{Z}}gank, Mirjam~Sepesy Mau{\v{c}}ec, Mitja Trojar, Jerneja~{\v{Z}}ganec Gros, Marko Bajec, Iztok~Lebar Bajec, and Simon Dobri{\v{s}}ek.
\newblock Strategies for managing time and costs in speech corpus creation: insights from the {Slovenian} {ARTUR} corpus.
\newblock {\em Language Resources and Evaluation}, 59(3):1899--1924, 2025.

\bibitem{mujadia2025ilgov}
Vandan Mujadia, Rao~B Ashwath, and Dipti~Misra Sharma.
\newblock Il-ilgov-2024: a translation benchmark for hindi-to-12 languages in the governance domain.
\newblock {\em Language Resources and Evaluation}, pages 1--22, 2025.

\bibitem{abdu2025empirical}
Fahad~J Abdu, Raed Mughaus, Shadi Abudalfa, Moataz Ahmed, and Ahmed Abdelali.
\newblock An empirical evaluation of arabic text formality transfer: a comparative study.
\newblock {\em Language Resources and Evaluation}, pages 1--61, 2025.

\bibitem{chakravarthi2025detecting}
Bharathi~Raja Chakravarthi, Saranya Rajiakodi, Rahul Ponnusamy, Bhuvaneswari Sivagnanam, Sara~Yogesh Thakare, and Sathiyaraj Thangasamy.
\newblock Detecting caste and migration hate speech in low-resource tamil language: Br chakravarthi et al.
\newblock {\em Language Resources and Evaluation}, pages 1--36, 2025.

\bibitem{SchickSchuetze2021}
Timo Schick and Hinrich Sch{\"u}tze.
\newblock Generating datasets with pretrained language models.
\newblock In {\em Proceedings of the 2021 Conference on Empirical Methods in Natural Language Processing (EMNLP)}, pages 6943--6951, Online and Punta Cana, Dominican Republic, 2021. Association for Computational Linguistics.

\bibitem{LampleEtAl2018}
Guillaume Lample, Alexis Conneau, Ludovic Denoyer, and Marc'Aurelio Ranzato.
\newblock Unsupervised machine translation using monolingual corpora only.
\newblock In {\em Proceedings of the International Conference on Learning Representations (ICLR)}, Vancouver, Canada, 2018.

\bibitem{DevlinEtAl2019}
Jacob Devlin, Ming-Wei Chang, Kenton Lee, and Kristina Toutanova.
\newblock {BERT}: Pre-training of deep bidirectional transformers for language understanding.
\newblock In {\em Proceedings of the 2019 Conference of the North American Chapter of the Association for Computational Linguistics: Human Language Technologies, Volume 1 (Long and Short Papers)}, pages 4171--4186, Minneapolis, Minnesota, 2019. Association for Computational Linguistics.

\bibitem{ConneauEtAl2020}
Alexis Conneau, Kartikay Khandelwal, Naman Goyal, Vishrav Chaudhary, Guillaume Wenzek, Francisco Guzm{\'a}n, Edouard Grave, Myle Ott, Luke Zettlemoyer, and Veselin Stoyanov.
\newblock Unsupervised cross-lingual representation learning at scale.
\newblock In {\em Proceedings of the 58th Annual Meeting of the Association for Computational Linguistics}, pages 8440--8451, Online, 2020. Association for Computational Linguistics.

\bibitem{CostaJussaEtAl2022}
Marta~R. Costa-jussà, James Cross, Onur Çelebi, Maha Elbayad, Kenneth Heafield, Kevin Heffernan, Elahe Kalbassi, Janice Lam, Daniel Licht, Jean Maillard, et~al.
\newblock No language left behind: Scaling human-centered machine translation, 2022.

\bibitem{BobojonovaEtAl2025}
Latofat Bobojonova, Arofat Akhundjanova, Phil~Sidney Ostheimer, and Sophie Fellenz.
\newblock {BBPOS}: {BERT}-based part-of-speech tagging for {Uzbek}.
\newblock In {\em Proceedings of the First Workshop on Language Models for Low-Resource Languages (LoResLM 2025)}, pages 287--293, Abu Dhabi, UAE, 2025. Association for Computational Linguistics.

\bibitem{AminEtAl2025}
Muhammad~Saad Amin, Xiao Zhang, Luca Anselma, Alessandro Mazzei, and Johan Bos.
\newblock Semantic processing for {Urdu}: corpus creation, parsing, and generation.
\newblock {\em Language Resources and Evaluation}, 59(3):2469--2500, 2025.

\bibitem{AungDras2025}
Kyaw~Htet Aung and Mark Dras.
\newblock {Myanmar} {XNLI}: building a dataset and exploring low-resource approaches to natural language inference with {Myanmar}.
\newblock {\em Language Resources and Evaluation}, 59(3):3267--3310, 2025.

\bibitem{GeteEtAl2025}
Harritxu Gete, Thierry Etchegoyhen, Gorka Labaka, Ander Corral, and Xabier Saralegi.
\newblock {TANDO+}: corpus and baselines for document-level machine translation in {Basque}--{Spanish} and {Basque}--{French}.
\newblock {\em Language Resources and Evaluation}, 2025.

\bibitem{ReichelEtAl2025}
Jaroslav Reichel and Vladim{\'i}r Benko.
\newblock Preservation of sentiment in machine translation of low-resource languages: a case study on {Slovak} movie subtitles.
\newblock {\em Language Resources and Evaluation}, 2025.

\bibitem{YazarEtAl2025}
Togay Yazar, Mucahid Kutlu, and {\.I}sa~Kerem Bay{\i}rl{\i}.
\newblock {Turkronicles}: diachronic resources for the fast evolving {Turkish} language.
\newblock {\em Language Resources and Evaluation}, 2025.
\newblock Published online: 2025.

\bibitem{HellwigBiagetti2025}
Oliver Hellwig and Erica Biagetti.
\newblock The {Sanskrit} {Sembank}.
\newblock {\em Language Resources and Evaluation}, 2025.
\newblock Published online: 2025.

\bibitem{PathakEtAl2025Bodo}
Dhrubajyoti Pathak, Sanjib Narzary, Sukumar Nandi, and Bidisha Som.
\newblock Part-of-speech tagger for {Bodo} language using deep learning approach.
\newblock {\em Natural Language Engineering}, 31(2):215--229, 2025.

\bibitem{Tham2020}
Medari Tham.
\newblock {NLP} tools for {Khasi}, a low resource language.
\newblock In {\em Proceedings of the 17th International Conference on Natural Language Processing (ICON): System Demonstrations}, pages 26--27, Patna, India, 2020. NLP Association of India (NLPAI).

\bibitem{TalukdarSarma2024}
Kuwali Talukdar and Shikhar~Kumar Sarma.
\newblock Enabling natural language processing and {AI} research in low-resource languages: Development and description of an {Assamese} {UPoS} tagged dataset.
\newblock {\em Journal of Electrical Systems}, 20(3):385--397, 2024.

\bibitem{PathakEtAl2022AsNER}
Dhrubajyoti Pathak, Sukumar Nandi, and Priyankoo Sarmah.
\newblock {AsNER}: Annotated dataset and baseline for {Assamese} named entity recognition.
\newblock In {\em Proceedings of the Thirteenth Language Resources and Evaluation Conference (LREC)}, pages 6571--6577, Marseille, France, 2022. European Language Resources Association.

\bibitem{SarmaPathak2023}
Surav Sarma and Nabankur Pathak.
\newblock Design and implementation of an {Assamese} language chatbot using neural networks.
\newblock {\em International Journal of Scientific Research in Computer Science and Engineering}, 11(6):13--18, 2023.

\bibitem{TamangBora2024}
Sagar Tamang and Dibya~Jyoti Bora.
\newblock Enhancing {Assamese} {NLP} capabilities: Introducing a centralized dataset repository, 2024.

\bibitem{NekotoEtAl2020}
Wilhelmina Nekoto, Vukosi Marivate, Tshinondiwa Matsila, Timi Fasubaa, Taiwo Fagbohungbe, Solomon~Oluwole Akinola, Shamsuddeen Muhammad, Salomon Kabiri-Haber, et~al.
\newblock Participatory research for low-resourced machine translation: A case study in {African} languages.
\newblock In {\em Findings of the Association for Computational Linguistics: EMNLP 2020}, pages 4184--4194, Online, 2020. Association for Computational Linguistics.

\bibitem{SennrichEtAl2016}
Rico Sennrich, Barry Haddow, and Alexandra Birch.
\newblock Improving neural machine translation models with monolingual data.
\newblock In {\em Proceedings of the 54th Annual Meeting of the Association for Computational Linguistics (Volume 1: Long Papers)}, pages 86--96, Berlin, Germany, 2016. Association for Computational Linguistics.

\bibitem{KartikEtAl2024}
{Kartik}, Sumanth Soni, Anoop Kunchukuttan, Tirthankar Chakraborty, and Md~Shad Akhtar.
\newblock Synthetic data generation and joint learning for robust code-mixed translation.
\newblock In {\em Proceedings of the 2024 Joint International Conference on Computational Linguistics, Language Resources and Evaluation (LREC-COLING)}, pages 4611--4622, Torino, Italy, 2024. ELRA and ICCL.

\bibitem{LancherosEtAl2025}
Brayan~Stiven Lancheros, Gloria Corpas~Pastor, and Ruslan Mitkov.
\newblock Data augmentation and transfer learning for cross-lingual named entity recognition in the biomedical domain.
\newblock {\em Language Resources and Evaluation}, 2025.

\bibitem{SaxenaEtAl2025}
Ashita Saxena, Dishank Aggarwal, Naveen Badathala, and Pushpak Bhattacharyya.
\newblock A framework to synthetically generate fine-grained hallucinated data.
\newblock {\em Language Resources and Evaluation}, pages 1--30, 2025.
\newblock Online First.

\bibitem{WinataEtAl2023}
Genta~Indra Winata, Alham~Fikri Aji, Zheng-Xin Yong, and Thamar Solorio.
\newblock The decades progress on code-switching research in {NLP}: A systematic survey on trends and challenges.
\newblock In {\em Findings of the Association for Computational Linguistics: ACL 2023}, pages 1496--1513, Toronto, Canada, 2023. Association for Computational Linguistics.

\bibitem{GuptaEtAl2021}
Abhirut Gupta, Aditya Vavre, and Sunita Sarawagi.
\newblock Training data augmentation for code-mixed translation.
\newblock In {\em Proceedings of the 2021 Conference of the North American Chapter of the Association for Computational Linguistics: Human Language Technologies}, pages 5760--5766, Online, 2021. Association for Computational Linguistics.

\bibitem{AguilarEtAl2020}
Gustavo Aguilar, Sudipta Kar, and Thamar Solorio.
\newblock {LinCE}: A centralized benchmark for linguistic code-switching evaluation.
\newblock In {\em Proceedings of the Twelfth Language Resources and Evaluation Conference}, pages 1803--1813, Marseille, France, 2020. European Language Resources Association.

\bibitem{ScholmanEtAl2025}
Merel C.~J. Scholman, Julie Hunter, Hiroyoshi Yamasaki, and Vera Demberg.
\newblock {DiscoNaija}: a discourse-annotated parallel {Nigerian} {Pidgin}--{English} corpus.
\newblock {\em Language Resources and Evaluation}, 2025.
\newblock Published online: 2025.

\bibitem{HuEtAl2022}
Edward~J. Hu, Yelong Shen, Phillip Wallis, Zeyuan Allen-Zhu, Yuanzhi Li, Shean Wang, Lu~Wang, and Weizhu Chen.
\newblock {LoRA}: Low-rank adaptation of large language models.
\newblock In {\em Proceedings of the International Conference on Learning Representations (ICLR)}, Online, 2022.

\bibitem{SuEtAl2024}
Tong Su, Xin Peng, Sarubi Thillainathan, David Guzm{\'a}n, Surangika Ranathunga, and En-Shiun Lee.
\newblock Unlocking parameter-efficient fine-tuning for low-resource language translation.
\newblock In {\em Findings of the Association for Computational Linguistics: NAACL 2024}, pages 4217--4225, Mexico City, Mexico, 2024. Association for Computational Linguistics.

\bibitem{MahdizadehSaniEtAl2025}
Soroush Mahdizadeh~Sani, Pegah Sadeghi, Tri-Thuan Vu, Yadollah Yaghoobzadeh, and Gholamreza Haffari.
\newblock Extending {LLMs} to new languages: A case study of {Llama} and {Persian} adaptation.
\newblock In {\em Proceedings of the 31st International Conference on Computational Linguistics (COLING 2025)}, Abu Dhabi, UAE, 2025. Association for Computational Linguistics.

\bibitem{AbbasiEtAl2023}
Mohammad~Amin Abbasi, Arash Ghafouri, Mahdi Firouzmandi, Hassan Naderi, and Behrouz Minaei-Bidgoli.
\newblock {PersianLLaMA}: Towards building first {Persian} large language model, 2023.

\bibitem{ScalviniEtAl2025}
Barbara Scalvini, Iben~Nyholm Debess, Annika Simonsen, and Hafsteinn Einarsson.
\newblock Rethinking low-resource {MT}: The surprising effectiveness of fine-tuned multilingual models in the {LLM} age.
\newblock In {\em Proceedings of the Joint 25th Nordic Conference on Computational Linguistics and 11th Baltic Conference on Human Language Technologies (NoDaLiDa/Baltic-HLT 2025)}, pages 609--621, Tallinn, Estonia, 2025.

\bibitem{NivreEtAl2020}
Joakim Nivre, Marie-Catherine de~Marneffe, Filip Ginter, Jan Hajič, Christopher~D. Manning, Sampo Pyysalo, Sebastian Schuster, Francis Tyers, and Daniel Zeman.
\newblock Universal dependencies v2: An evergrowing multilingual treebank collection.
\newblock In {\em Proceedings of the Twelfth Language Resources and Evaluation Conference}, pages 4034--4043, Marseille, France, 2020. European Language Resources Association.

\bibitem{AlzettaEtAl2025}
Chiara Alzetta, Alessio Miaschi, Felice Dell'Orletta, Giulia Venturi, and Simonetta Montemagni.
\newblock Parallel trees: a novel resource with aligned dependency and constituency syntactic representations.
\newblock {\em Language Resources and Evaluation}, 2025.

\bibitem{TjongKimSangVeenstra1999}
Erik~F. Tjong Kim~Sang and Jorn Veenstra.
\newblock Representing text chunks.
\newblock In {\em Proceedings of the Ninth Conference of the European Chapter of the Association for Computational Linguistics}, pages 173--179, Bergen, Norway, 1999. Association for Computational Linguistics.

\bibitem{Cohen1960}
Jacob Cohen.
\newblock A coefficient of agreement for nominal scales.
\newblock {\em Educational and Psychological Measurement}, 20(1):37--46, 1960.

\bibitem{LiEtAl2022}
Hongwei Li, Hongyan Mao, and Jingzi Wang.
\newblock Part-of-speech tagging with rule-based data preprocessing and transformer.
\newblock {\em Electronics}, 11(1):56, 2022.

\bibitem{LoshchilovHutter2019}
Ilya Loshchilov and Frank Hutter.
\newblock Decoupled weight decay regularization.
\newblock In {\em Proceedings of the International Conference on Learning Representations (ICLR)}, New Orleans, Louisiana, 2019.

\bibitem{seqeval}
Hiroki Nakayama.
\newblock {seqeval}: A python framework for sequence labeling evaluation, 2018.
\newblock Software available from https://github.com/chakki-works/seqeval.

\bibitem{BrownEtAl1990}
Peter~F. Brown, John Cocke, Stephen~A. Della~Pietra, Vincent~J. Della~Pietra, Fredrick Jelinek, John~D. Lafferty, Robert~L. Mercer, and Paul~S. Roossin.
\newblock A statistical approach to machine translation.
\newblock {\em Computational Linguistics}, 16(2):79--85, 1990.

\bibitem{Lin2004ROUGE}
Chin-Yew Lin.
\newblock {ROUGE}: A package for automatic evaluation of summaries.
\newblock In {\em Proceedings of the ACL-04 Workshop on Text Summarization Branches Out}, pages 74--81, Barcelona, Spain, 2004. Association for Computational Linguistics.

\bibitem{PapineniEtAl2002BLEU}
Kishore Papineni, Salim Roukos, Todd Ward, and Wei-Jing Zhu.
\newblock {BLEU}: a method for automatic evaluation of machine translation.
\newblock In {\em Proceedings of the 40th Annual Meeting of the Association for Computational Linguistics}, pages 311--318, Philadelphia, PA, 2002. Association for Computational Linguistics.

\bibitem{Popovic2017chrFpp}
Maja Popovi{\'c}.
\newblock chrf++: words helping character n-grams.
\newblock In {\em Proceedings of the Second Conference on Machine Translation, Volume 2: Shared Task Papers}, pages 612--618, Copenhagen, Denmark, 2017. Association for Computational Linguistics.

\bibitem{ReiEtAl2020}
Ricardo Rei, Craig Stewart, Ana~C Farinha, and Alon Lavie.
\newblock {COMET}: A neural framework for {MT} evaluation.
\newblock In {\em Proceedings of the 2020 Conference on Empirical Methods in Natural Language Processing (EMNLP)}, pages 2685--2702, Online, 2020. Association for Computational Linguistics.

\end{thebibliography}

\end{document}